\documentclass[final]{cvpr}

\usepackage{times}
\usepackage{epsfig}
\usepackage{graphicx}
\usepackage{amsmath}
\usepackage{amssymb}
\usepackage{multirow}
\usepackage{algorithm}
\usepackage{algorithmic}
\usepackage{diagbox}
\usepackage{booktabs}

\usepackage[pagebackref=true,breaklinks=true,colorlinks,bookmarks=false]{hyperref}

\def\cvprPaperID{6009} 
\def\confYear{CVPR 2021}

\begin{document}

\title{An Eye for an Eye: Defending against Gradient-based Attacks with Gradients }
\author{Hanbin Hong\\
Illinois Institute of Technology\\
{\tt\footnotesize hhong4@hawk.iit.edu}
\and
Yuan Hong\\
Illinois Institue of Technology\\
{\tt\footnotesize yuan.hong@iit.edu}
\and
Yu Kong\\
Rochester Institute of Technology\\
{\tt\footnotesize yu.kong@rit.edu}
}

\maketitle

\begin{abstract}
    Deep learning models have been shown to be vulnerable to adversarial attacks. In particular, gradient-based attacks have demonstrated high success rates recently. The gradient measures how each image pixel affects the model output, which contains critical information for generating malicious perturbations. In this paper, we show that the gradients can also be exploited as a powerful weapon to defend against adversarial attacks. By 
   using both gradient maps and adversarial images as inputs, we propose a Two-stream Restoration Network (\textbf{TRN}) to restore the adversarial images. To optimally restore the perturbed images with two streams of inputs, a Gradient Map Estimation Mechanism is proposed to estimate the gradients of adversarial images, and a Fusion Block is designed in TRN to explore and fuse the information in two streams. Once trained, our TRN can defend against a wide range of attack methods without significantly degrading the performance of benign inputs. Also, our method is generalizable, scalable, and hard to bypass. Experimental results on CIFAR10, SVHN, and Fashion MNIST demonstrate that our method outperforms state-of-the-art defense methods.  
\end{abstract}

\begin{figure}
    \centering
    \includegraphics[scale=0.34]{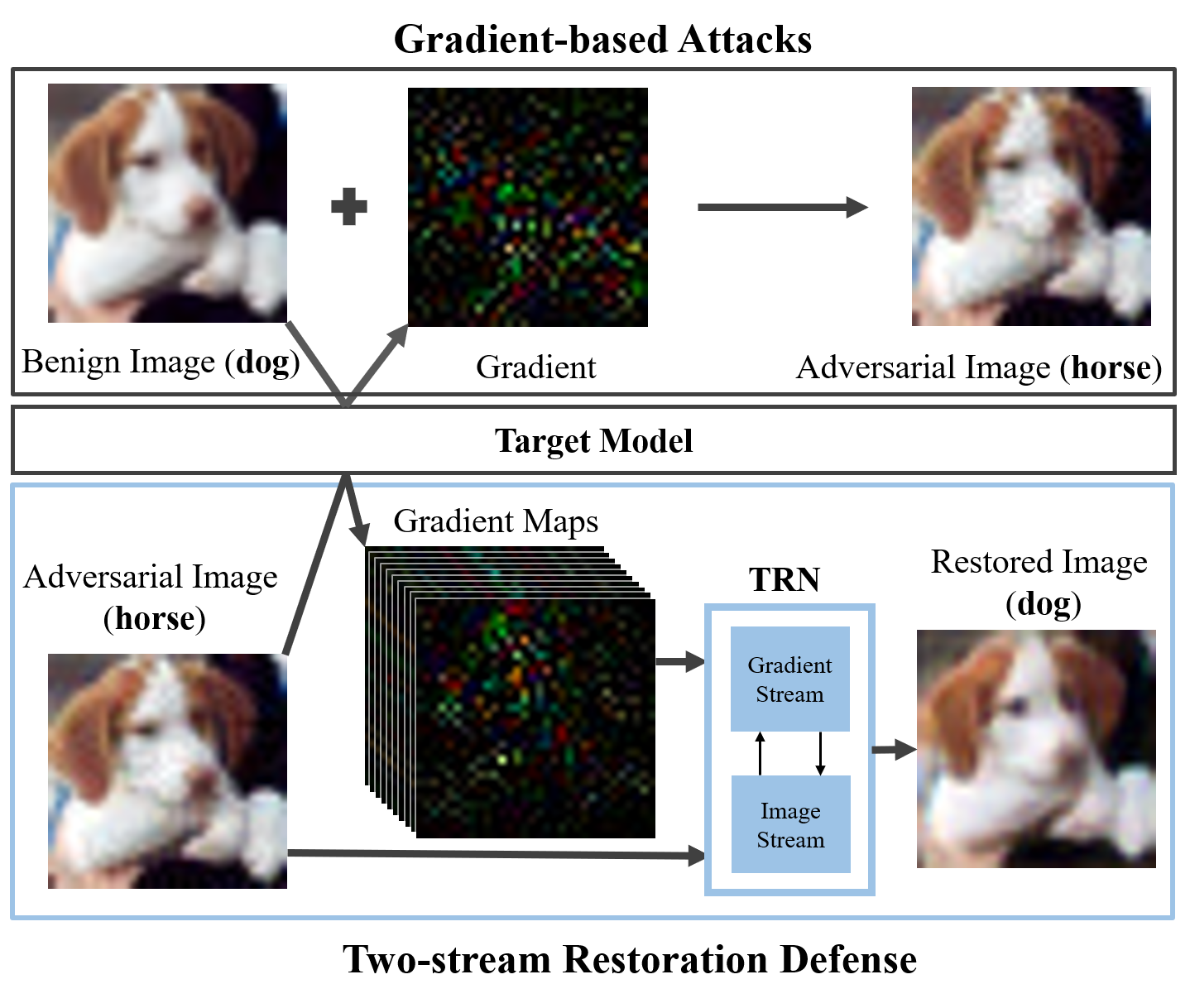}
    \caption{Gradient-based attacks can craft adversarial images to fool the target model, e.g., changing the predicted class from dog to horse. In this paper, we show that the gradient can also be exploited to benefit defense methods. Incorporating both image and gradient streams, our Two-stream Restoration Defense can destroy fatal adversarial patterns, and restore the image representation.}
    \label{fig:Idea}
\end{figure}

\section{Introduction}

    Adversarial attacks aim to fool the deep neural networks to provide incorrect outputs \cite{biggio2013evasion,carlini2019evaluating}. The distance between the outputs of clean images and adversarial images is maximized to construct adversarial images. Following this path, a series of white-box attacks, including Fast Gradient Sign Method (FGSM) \cite{goodfellow2014explaining}, Basic Iterative Method (BIM) \cite{kurakin2016adversarial}, and Projected Gradient Descent (PGD) \cite{madry2017towards}, craft adversarial images according to the gradients of the loss function with respect to the input. The gradient measures how small changes at each input pixel affect the model outputs \cite{chan2020thinks}. With access to the gradient, adversaries are able to craft adversarial examples along the direction of loss 
   ascending. Thus, gradient-based attacks achieve high success rates in attacking deep learning models \cite{dong2020benchmarking,tramer2017ensemble,kannan2018adversarial,zhang2019theoretically,naseer2020self}.

    
    Many proposed defense methods are able to defend against gradient-based attacks. The robust training based methods \cite{madry2017towards, goodfellow2014explaining, tramer2017ensemble,zhang2019theoretically,hein2017formal,ross2017improving,yan2018deep,lecuyer2019certified} aim to train a robust target model either by including adversarial examples in training data or by regularization. The input transformation based methods \cite{guo2017countering,jia2019comdefend,xie2017mitigating,liao2018defense,naseer2020self,yuan2020ensemble,samangouei2018defense,song2017pixeldefend} modify the image examples to invalidate the adversarial effect before feeding them to the target model. The randomization based methods \cite{xie2017mitigating,pang2019mixup,cohen2019certified,dhillon2018stochastic,liu2018towards,liu2018adv} reduce the adversarial effects by randomizing the inputs or model parameters.  However, these methods can hardly recover the performance of the target model to the original level. Comparing to the attack methods which explore the gradient to craft adversarial images, defense methods unequally ignore the critical information about how each pixel changes of inputs affect the model outputs. The information provided by adversarial images is insufficient for defense methods to mitigate the adversarial effects since images contain the zero-order information of the target model while the gradient contains the first-order information of the target model \cite{madry2017towards}. Therefore, we leverage the adversarial images and their gradients to defend against gradient-based attacks. The gradients of adversarial images measure how small changes at each pixel of adversarial images affect the model outputs. This enables our method to incorporate extra information beyond the adversarial images and further provide a more robust defense.

    In this paper, we propose a Two-stream Restoration Network (\textbf{TRN}) to explore and fuse the information in adversarial images and its gradients, and further restore the adversarial images to a benign state (see Figure \ref{fig:Idea}). To utilize the gradient on the defense side, we addressed two challenges. The first one is to obtain the gradient. The gradients are available in attack methods because white-box attacks know both the network parameters and the label of input images. Adversaries can easily compute the gradient through backward propagation \cite{papernot2016technical,rauber2017foolbox,art2018,dong2020benchmarking}. However, the gradient is hard to obtain on the defense side since defenders do not have access to the label of original images. To solve this problem, we design a Gradient Map Estimation Mechanism (\textbf{GMEM}) to derive a set of estimated gradient maps. Another challenge is how to integrate information in the image stream and gradient stream since the gradient and the adversarial images are from different spaces. To bridge the features of two streams, we design a Fusion Block (\textbf{FB}) in TRN to fuse these two streams. The Fusion Block can be simply concatenated to construct a deeper network. In this case, we are able to train a TRN with different network depths to handle target models with different complexities.
    
      
    
    Therefore, our contributions can be summarized as:
    \begin{itemize}
    \item We show that the gradients of adversarial images can be exploited to defend against gradient-based attacks. Considering both adversarial images and the corresponding gradients, we propose a Two-stream Restoration Network for defense. To our best knowledge, this is the first attempt to leverage the gradients of adversarial images to defend against adversarial attacks.
    \item To solve two major challenges in using gradients on the defense side, we propose a novel Gradient Map Estimation Mechanism to estimate the gradients of adversarial images. Also, a Fusion Block is designed to fuse the image stream and gradient stream.
    \item Our new method outperforms the state-of-the-art defense methods on CIFAR10, SVHN, and Fashion MNIST. Experiments also demonstrate that our defense is generalizable, scalable, and hard to bypass.

\end{itemize}

\section{Related Work}

\textbf{Attack}: Gradient-based attacks \cite{goodfellow2014explaining,kurakin2016adversarial,madry2017towards} exploit the benefits of back-propagation to generate gradients and further craft the adversarial images. Specifically, the FGSM \cite{goodfellow2014explaining} is a single-step attack that transforms the gradient of the loss function with a sign function to generate the perturbation. 
Goodfellow \textit{et al.} \cite{kurakin2016adversarial} provide the Basic Iterative Method to generate perturbations in an iterative fashion. Perturbations are generated based on the gradients of the loss function with respect to the adversarial images in the last iteration. Similarly, the PGD \cite{madry2017towards} was proposed to improve the BIM by randomly initializing the perturbation within the perturbation budget. It also applies random restarts after certain iterations. MIM \cite{dong2018boosting} uses momentum terms to prevent iterative optimization from staying in poor local maxima. FFGSM \cite{wong2020fast} improves FGSM in adversarial training to make the performance comparable to PGD while reducing the training time. Besides the attack that directly uses gradients to compute adversarial images, the C{\&}W attack \cite{carlini2016defensive} crafts adversarial examples by minimizing perturbations. Moreover, 
Carlini \textit{et al.} \cite{athalye2018obfuscated} first train a network to approximate the input transformation, then compute the gradient based on the approximation network.

\textbf{Defense}: Many input transformation based methods \cite{guo2017countering,xie2017mitigating,liao2018defense,samangouei2018defense} are proposed to eliminate the adversarial patterns on images. Guo \textit{et al.} \cite{guo2017countering} transform images with bit-depth reduction, JPEG compression, total variance minimization, and image quilting to process images before feeding them to networks. Xie \textit{et al.} \cite{xie2017mitigating} use random resizing and random padding to modify inputs to mitigate the adversarial effect. HGD \cite{liao2018defense} proposes a high-level representation guided denoiser to process images to defend against attacks. Defense-GAN \cite{samangouei2018defense} trains a Generative Adversarial Network to transform the adversarial images into clean examples. Recent methods \cite{naseer2020self,yuan2020ensemble,madry2017towards} leverage adversarial training to design pre-processing networks, which can successfully defend against a variety of attacks. 
Other methods \cite{lecuyer2019certified,cohen2019certified} leverage differential privacy or randomization to make the model more robust. However, all the above defenses solely rely on the adversarial images as inputs. The gradients of adversarial images, which contain first-order information of target models, are ignored in these methods. Thus, these methods fail to provide optimal defense performance to the target model. 


\begin{figure}
    \centering
    \includegraphics[scale=0.33]{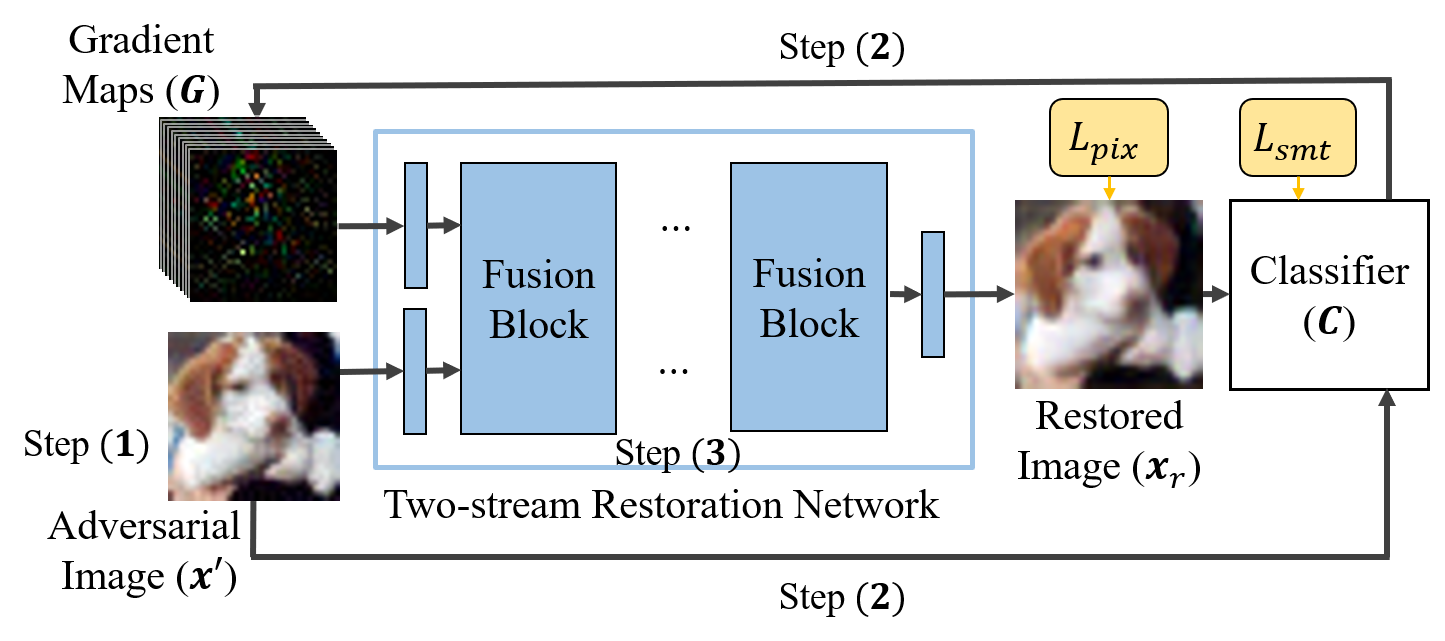}
    \caption{Two-stream Restoration Defense. To achieve the optimal robustness against gradient-based attacks, we adopt adversarial training to train a Two-stream Network as the pre-processing network. The training process iterates the following three steps: \textbf{(1)} Generating adversarial images using gradient-based attack algorithms. \textbf{(2)} Computing the gradient maps of adversarial images using GMEM. \textbf{(3)} Optimizing the TRN to restore the adversarial images with gradient maps and adversarial images. }
    \label{fig2:Framework}
\end{figure}
\section{Two-stream Restoration Network}

Our TRN serves as a pre-processing network to restore the adversarial images (see Figure \ref{fig2:Framework}). Different from other defense methods \cite{naseer2020self,yuan2020ensemble}, our TRN incorporates not only the adversarial images but also the gradients of adversarial images to defend against gradient-based attacks. Recall that the gradients measure how small pixel changes affect the model output. Considering the gradient maps as the second stream, our defense can outperform single-stream defense methods (See experimental results in Table \ref{tab:table1} and \ref{tab:table4}). 

\textbf{Adversarial Training}: We adopt adversarial training \cite{madry2017towards} to optimize our TRN. Given the benign image $\textbf{x}$ and label $\textbf{y}$, we first generate the adversarial images $\textbf{x}'$: $\textbf{x}'=\textbf{x}+\delta$ by maximizing the adversarial loss $\mathcal{L}_{adv}$.


\begin{equation}\label{eq2}
\delta=\mathop{\arg\max}_{\delta}\mathcal{L}_{adv}(\mathcal{C}(\textbf{x}'),\textbf{y}), ||\delta||_{\infty} \leq \epsilon
\end{equation} where $\delta$ is the perturbation. $\epsilon$ is the perturbation budget in $\ell_\infty$ norm. $\mathcal{C}(\cdot)$ denotes the target model. Many gradient-based attack algorithms can be adopt to solve the maximization problem. For example, the Fast Gradient Sign Method \cite{goodfellow2014explaining} solves this optimization by $\delta=\epsilon sgn(\nabla_{\textbf{x}}{\mathcal{L}_{adv}})$, where $sgn(\cdot)$ denotes the sign function. Usually, the adversarial loss for classification task is the cross-entropy loss.

Once the adversarial images are crafted, we then compute the gradient maps $\textbf{G}$ of adversarial images using GMEM (see Section 3.1). The restored image $\textbf{x}_r$ will be generated using our TRN.

\begin{figure}
    \centering
    \includegraphics[scale=0.32]{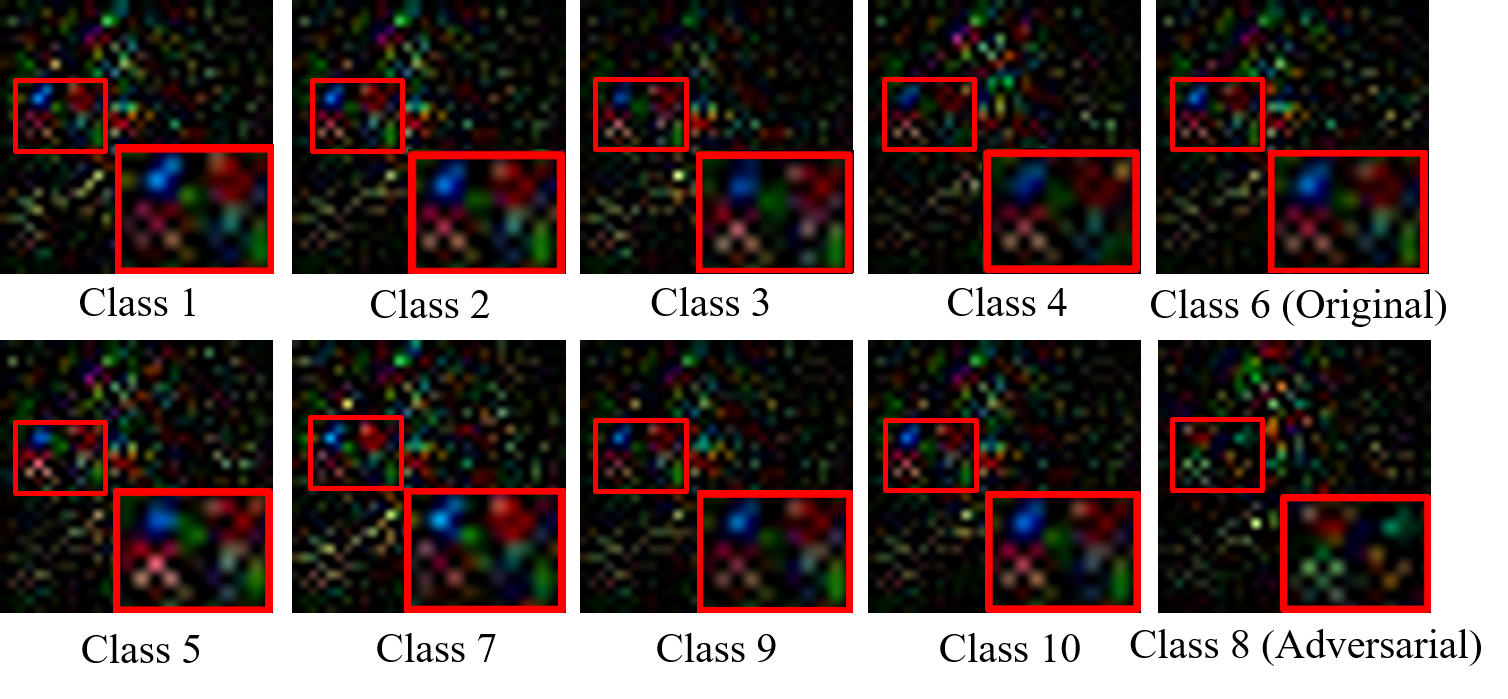}
    \caption{Visualization of Gradient Maps on CIFAR10 dataset. The red box is zoomed in for details. The spatial distributions of gradient maps are similar, while the colors of the gradient map in the adversarial class (Class 8) are different from those in other classes. The color difference results from the difference of adversarial losses since the adversarial image is classified into the adversarial class}
    \label{fig:gradient_maps}
\end{figure}

\begin{algorithm}
\small
\caption{Two-stream Restoration Defense}
\label{alg:A}
\renewcommand{\algorithmicrequire}{\textbf{Input:}}
\renewcommand{\algorithmicensure}{\textbf{Output:}}
\begin{algorithmic}[1]
\REQUIRE Training data $D$, Two-stream Restoration Network $TRN$, classifier $\mathcal{C}$, perturbation budget $\epsilon$ and loss criteria $\mathcal{L}$
\ENSURE Randomly initialize $TRN$
\REPEAT
\STATE Sample mini-batch of data $\textbf{x}$ from the training set.
\STATE Find adversarial images $\textbf{x}'$ by Eq. \ref{eq2}.
\STATE Generating gradient maps $\textbf{G}$ using Eq. \ref{eq6}.
\STATE Forward-pass $\textbf{G}$ and $\textbf{x}'$ through $TRN$ (Eq. \ref{eq3}) to generate $\textbf{x}_r$ and calculate $\mathcal{L}_{pix}$.
\STATE Forward-pass $\textbf{x}_r$ through $\mathcal{C}$ and calculate $\mathcal{L}_{smt}$.
\STATE Back-pass and update $\theta_{TRN}$ to minimize $\mathcal{L}_{pix}$ and $\mathcal{L}_{smt}$.
\UNTIL{$TRN$ converges.}
\end{algorithmic}
\end{algorithm}

\begin{equation}\label{eq3}
    \textbf{x}_{r}=TRN(\textbf{G},\textbf{x}',\theta_{TRN})
\end{equation} where $\theta_{TRN}$ denotes the network parameters of TRN. Different from other input transformation based defenses \cite{naseer2020self,yuan2020ensemble,mustafa2019image}, we leverage both image and gradient streams to process the adversarial image, which can provide a more robust defense. To obtain an optimal TRN that can restore the adversarial images in both pixel space and semantic space, we minimize the empirical risk $\rho$ given the dataset $D$: 

\begin{equation}\label{eq4}
    \min_{\theta_{TRN}} \rho(\theta_{TRN})
\end{equation}
\begin{equation}\label{eq5}
    \rho(\theta_{TRN})=\mathbb{E}_{(\textbf{x},\textbf{y}){\thicksim}D}[\mathcal{L}_{pix}(\textbf{x},\textbf{x}_r)+\mathcal{L}_{smt}(\mathcal{C}(\textbf{x}_r),\textbf{y})]
\end{equation}

Note that we have two loss functions: pixel loss $\mathcal{L}_{pix}$ and semantic loss $\mathcal{L}_{smt}$. The pixel loss is proposed to restore the pixel domain of adversarial images, which is the $\ell_2$ distance between restored image $\textbf{x}_r$ and benign image $\textbf{x}$. The semantic loss aims to revert the performance of adversarial images on the classifier, which is the cross-entropy loss of classifier outputs $\mathcal{C}(\textbf{x}_r)$ and the label $\textbf{y}$. The detailed procedures are summarized in Algorithm \ref{alg:A}.

\begin{figure*}
    \centering
    \includegraphics[scale=0.45]{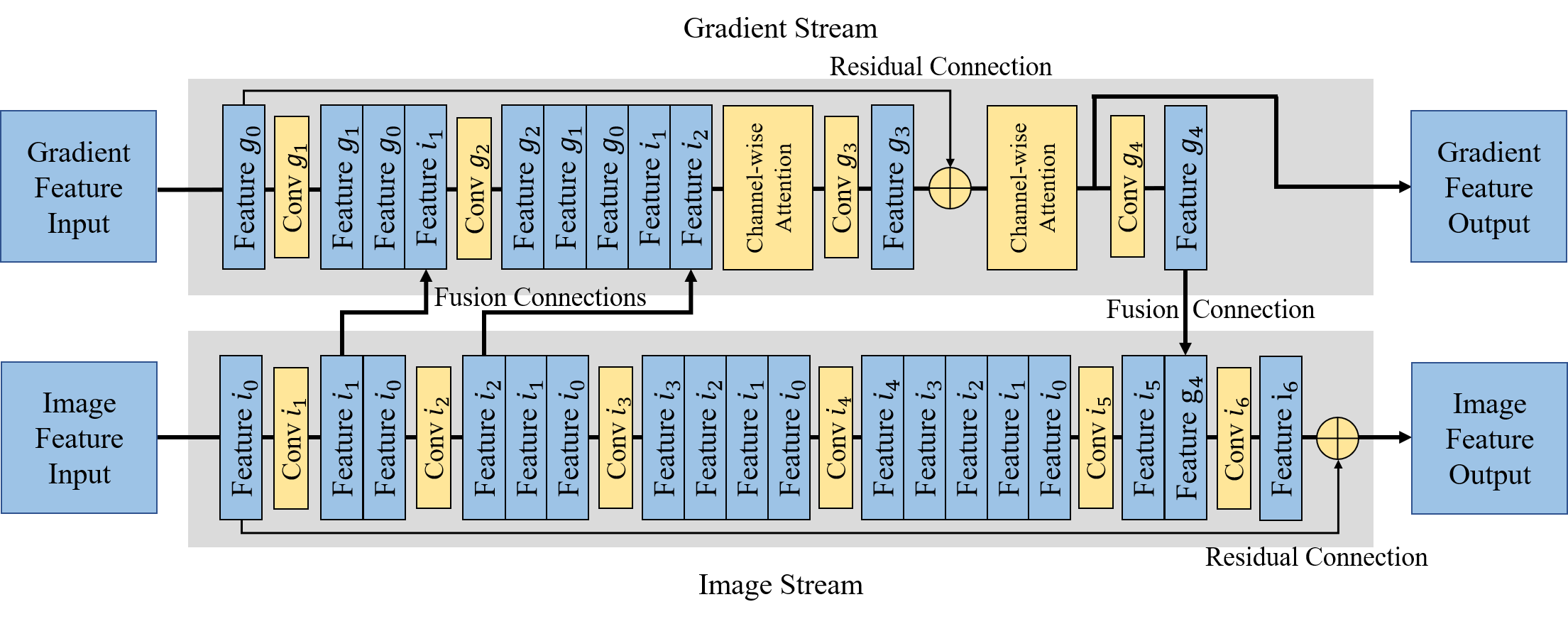}
    \caption{Fusion Block. The Fusion Block contains image stream and gradient stream. With the connections between two streams, the Fusion Block learns to fuse the information from two streams to restore images.}
    \label{fig2:fusion block}\vspace{-0.1in}
\end{figure*}

\subsection{Gradient Map Estimation Mechanism}

To incorporate the gradients of adversarial images for further restoration, we need to first compute the gradients. However, on the defense side, the gradients of adversarial images are unavailable since defenders do not have access to the original label of adversarial images. Thus, to this end, we propose a Gradient Map Estimation Mechanism for estimating the gradients.

For gradient-based attacks, the gradient is commonly computed according to $\nabla_\textbf{x}\mathcal{L}_{adv}(\mathcal{C}(\textbf{x}),\textbf{y})$ \cite{madry2017towards,goodfellow2014explaining,kurakin2016adversarial}, which requires the ground truth label $\textbf{y}$. For gradients of adversarial images $\nabla_{\textbf{x}'}\mathcal{L}_{adv}(\mathcal{C}(\textbf{x}'),\textbf{y})$, the ground truth label $\textbf{y} $ is unknown, but $\textbf{y}$ should be one of the possible labels $\{y_i\}, i=1,2,3 ..., n$, since the sample should belong to one class in a dataset. Thus, for each possible label $\textbf{y}_i$, we compute the gradient of adversarial images. In this way, we generate the gradient maps as below: 

\begin{equation}\label{eq6}
    \textbf{G}=\{\nabla_{\textbf{x}'}\mathcal{L}(\textbf{x}',\textbf{y}_i)\}, i=1,2,3,...n
\end{equation} where $y_i$ denotes the possible label in class $i$, $n$ denotes the total classes of the dataset. The gradient maps $\textbf{G}$ are considered as an estimator of the gradient computed by the real label. In our TRN, the gradient maps of all classes are then concatenated along the channel dimension. 

In Figure \ref{fig:gradient_maps}, we visualize the gradient maps of one adversarial image. These gradients measure how each pixel of the adversarial image contributes to the model outputs. In this case, our TRN can focus on the critical pixels that strongly affect the model outputs to provide more precise restoration. By comparing the area in red boxes, we find that the gradient maps generated using possible labels share similar spatial patterns, while the gradient map in the adversarial class share different colors with gradient maps in other classes. The similarity of spatial patterns results from the same input, target model, and loss function, while the color difference is due to different loss values. Since the image is crafted to be in the adversarial class, the loss value for the adversarial class is low, while the loss values for other classes are high.

\begin{table*}
\caption{Performance (classification accuracy after defense) of our method against adversarial attacks on CIFAR10 ($\epsilon=8/255$) , SVHN ($\epsilon=12/255$) and Fashion MNIST ($\epsilon=8/255$). Our method significantly outperforms the state-of-the-art defense by a large margin.}
\centering
\begin{tabular}{c|l|c| c c c c c c c}
\toprule[1.5pt]
                &Defense Methods                                 & No attack     & PGD \cite{madry2017towards}       & BIM \cite{kurakin2016adversarial}          & FGSM \cite{goodfellow2014explaining}          & MIM \cite{dong2018boosting}       & FFGSM \cite{wong2020fast}\\ \midrule[1pt]
\multirow{10}*{\rotatebox{90}{CIFAR10 \cite{krizhevsky2009learning}}}         &No defense     & 95.34         &0.01       & 0.01         & 37.60         & 0.08      & 33.02       \\ \cline{2-8}
                &JPEG \cite{guo2017countering}                                   & 76.91         &67.16      & 64.6         & 59.27         & 52.75     & 63.11        \\
                &Label Smoothing \cite{warde201611}                        & 92.93         &36.69      & 37.9          &57.62          &45.60       & 57.27     \\ 
                &Feature Squeezing \cite{xu2017feature}                      & 27.13         & 23.55     & 22.82         & 23.41         &21.80       & 24.18         \\ 
                &R{\&}P \cite{xie2017mitigating}                                 & 91.49          &60.09      &48.16          & 56.80          & 27.79     &56.84    \\ 
                &SR \cite{mustafa2019image}                                     & 82.31       &68.84      &65.00          &57.24          &45.07      &61.54     \\
                &NRP \cite{naseer2020self}                                    &89.96           &78.77      &72.67          &72.86          & 61.18     &75.91         \\
                &Adv. Training \cite{madry2017towards}                          & 76.48          &44.16      &38.66          &42.49          & 40.46     &51.06     \\ \cline{2-8}
                &Ours                                           & \textbf{94.37}          &\textbf{86.96}      &\textbf{90.01}          & \textbf{87.10}         &\textbf{91.90}     & \textbf{87.51}    \\ \midrule[1pt]
                
\multirow{10}*{\rotatebox{90}{SVHN \cite{netzer2011reading}}}         &No defense        & 97.11         &0.16       & 0.15         & 37.25         & 1.40      & 39.65       \\ \cline{2-8}
                &JPEG \cite{guo2017countering}                                   & 96.81         &1.02      & 0.70         & 39.32        & 3.70     & 43.91        \\
                &Label Smoothing \cite{warde201611}                        & 96.66         &4.67      & 3.79          &50.30         &17.71       &57.62     \\ 
                &Feature Squeezing \cite{xu2017feature}                      & 60.88         & 36.29     & 34.36         & 43.13         &31.76      & 46.22         \\ 
                &R{\&}P \cite{xie2017mitigating}                                 & 94.42          &28.00     &24.09          & 47.64         & 18.78     &53.39    \\ 
                &SR \cite{mustafa2019image}                                     & 95.59      &55.12      &46.13          & 51.88         & 28.07     & 59.39    \\
                &NRP \cite{naseer2020self}                                    & 92.74          & 89.24     &   74.38         &  89.47        &  63.41    & 82.27        \\
                &Adv. Training \cite{madry2017towards}                       &  95.44        &     35.8       &   35.35   &   78.7       &     51.01     &  61.71    \\ \cline{2-8}
                &Ours                                           & \textbf{97.12}          &\textbf{95.34}      &\textbf{95.70}          & \textbf{98.13}         &\textbf{98.89}     & \textbf{98.41}    \\ \midrule[1pt]
                
\multirow{10}*{\rotatebox{90}{Fashion MNIST \cite{xiao2017fashion}}}  &No defense      & 95.08         &0.00       & 0.01          & 55.32         & 0.02      & 41.48       \\ \cline{2-8}
                &JPEG \cite{guo2017countering}                  & 90.87         &71.82      & 68.97         & 66.99         & 60.51     & 68.89        \\
                &Label Smoothing \cite{warde201611}             & 90.61         &11.58      &19.33          & 51.42         & 25.93     & 46.58     \\ 
                &Feature Squeezing \cite{xu2017feature}         & 60.95         & 59.58     & 59.21         & 59.73         & 58.88     & 60.12         \\ 
                &R{\&}P \cite{xie2017mitigating}                & 90.37         & 39.05     & 35.48         & 60.37         & 32.39     & 57.56        \\ 
                &SR \cite{mustafa2019image}                     &  92.92        & 66.66     & 59.89      & 67.97         & 45.37     & 66.37    \\
                &NRP \cite{naseer2020self}                      &  92.69        & 84.73     & 77.84         & 78.21         & 71.37     & 82.09        \\
                &Adv. Training \cite{madry2017towards}          &   90.56       &    86.15       &  85.22   &     85.65     &    85.35      &    87.01            \\ \cline{2-8}
                &Ours                                           & \textbf{95.08 }         &\textbf{93.11}    &\textbf{92.75}          & \textbf{95.12}       &\textbf{96.42}     & \textbf{95.12}   \\ \bottomrule[1.5pt]
\end{tabular}
\label{tab:table1}
\end{table*}

\subsection{Fusion Block (FB)}

Our TRN takes both image stream and gradient stream as inputs to restore adversarial images. Both two streams are related to each other but they are from different spaces, in which the value distributions are quite different. How to fuse these two streams to jointly guide the restoration is a challenge. To address this challenge, we design the Fusion Block in TRN to explore and fuse the information from two streams. The architecture is shown in Figure \ref{fig2:fusion block}. 

\textbf{Residual Connection}: The nature of residual learning \cite{he2016deep} matches the goal of adversarial image restoration, which is to learn the benign perturbation adding to the adversarial images. Therefore, in the image stream, through the $6$ convolution layers, we learn the benign perturbation, then with the residual connection after convolution layer $i_6$, we add the benign perturbation to the adversarial image to obtain the restored image. The goal of the gradient stream is to help learn the benign perturbation since the natures of the gradient and perturbation are more close (perturbations are constructed from gradient in attack methods). Similarly, through the residual connection in the gradient stream, gradient stream learns to refine the gradient to provide more accurate guidance on generating perturbation.

\textbf{Dense Connectivity}: Motivated by Huang \textit{et al.} \cite{huang2017densely}, we use densely connected convolution in both streams, in which each layer takes all preceding feature-maps as input. In this way, the convolution layers can extract useful information from different level's features. Also, FBs can be concatenated to construct deeper networks to provide more accurate restoration without gradient vanishing.

\textbf{Fusion Connection}: Instead of fusing two streams into one stream \cite{Simonyan2014Two}, we use one stream to guide the other stream. First, by inserting features from image stream to gradient stream after the first two layers, we use the image features from shallow layers, which contains undistorted information, to guide the gradient refining. Second, as aforementioned, we use gradient stream to help learn the benign perturbation, so we concatenate the image features $i_5$ and the gradient features $g_4$ to learn the benign perturbation.

\textbf{Channel-wise Attention}: In the gradient stream, gradient maps and the features from two streams are all concatenated along the channel dimension. However, the contributions from gradient features and image features are different. Also, the gradient maps computed with different labels may have different contributions. Thus, we adopt the Channel-wise Attention (CA) \cite{hu2018squeeze} to re-weigh the feature in different channels. The first and second CA are used to re-weigh the features from different streams and the gradient map generated with different labels, respectively.

\section{Experiments}

In this section, we evaluate our TRN defense comprehensively. Our method is compared to $7$ state-of-the-art defense methods against $5$ gradient-based attacks on $3$ datasets. We evaluate our method in three aspects: (1) The defense performance against attacks. (2) The generalizability across attacks. (3) Robustness against bypass attacks.

\subsection{Experimental Setting}

The evaluation is based on three datasets including CIFAR-10 \cite{krizhevsky2009learning}, SVHN (Street View House Number) \cite{netzer2011reading} and Fashion MNIST \cite{xiao2017fashion}. The CIFAR-10 dataset has $60,000$ images in 10 classes. The SVHN dataset contains 10 classes of $99,289$ street view images in total.. The Fashion MNIST contains $60,000$ gray-scale images in 10 classes. Image sizes in CIFAR10 and SVHN datasets are $32\times32$. We also resize the images of Fashion MNIST to $32\times32$ from the original $28\times28$. 

For a fair comparison, we test all the attack and defense methods on the same target model for each dataset. We use a ResNet-18 \cite{he2016deep} as the target model.  All the methods are implemented with PyTorch \cite{paszke2019pytorch}. In our method, we use Adam as the optimizer with an initial learning rate of $1\times 10^{-3}$ and reduce the learning rate by a factor of $10$ every $30$ epochs. In the training, besides the adversarial images and the gradient maps, we include the benign images and the corresponding gradient maps to train the TRN.

\begin{table*}
\caption{Generalizability across different attacks on CIFAR10, SVHN, and Fashion MNIST (Fashion). We compare our method (Right) with adversarial training \cite{madry2017towards}  (Left). Methods are trained with attacks in the second column and test on the attack in the first row. We show the classification accuracy after the defense. Variation is the difference between maximum accuracy and minimum accuracy. Our method is generalizable across different attacks.}
\centering
\begin{tabular}{c| c | c c c c c c}
\toprule[1.5pt]
\multicolumn{2}{c|}{}    &  PGD &  BIM &  FGSM &  MIM &  FFGSM & Variation\\ \midrule[1pt]
    \multirow{3}{*}[-2.2ex]{\rotatebox{90}{\footnotesize CIFAR10 \cite{krizhevsky2009learning}}}     & PGD &43.04 / 89.72            &37.33 / 87.33            &41.85 / 73.44          &39.36 / 75.26         &49.87 / 79.74 &      12.54 / 16.28\\ 
                                                                                        & BIM &45.29 / 85.18           &40.04 / 89.67            &43.17 / 78.01          &41.41 / 84.62          &50.72 / 80.77     &   10.68 / 11.66 \\ 
                                                                                        & FGSM&2.39 / 87.88          &2.24 / 88.61            &74.29 / 87.87          &2.60 / 88.25          &90.67 / 87.77         &  88.43 / 0.84\\
                                                                                        & MIM &44.16 / 86.96           &38.66 / 90.01           &42.49 / 87.10           &40.46 / 91.90         &51.06 / 87.51      &     12.40 / 4.94\\
                                                                                        & FFGSM&39.15 / 88.66           &33.12 / 88.63            &38.79 / 86.97         &35.32 / 86.65          &47.72 / 88.80     &      14.60 / 2.15 \\\midrule[1pt]
                                                            
    \multirow{3}{*}[-2.2ex]{\rotatebox{90}{\footnotesize SVHN \cite{netzer2011reading}}}         & PGD &52.83 / 97.54            &10.58 / 95.52          &69.62 /  98.66          &4.61 / 98.66         &85.28 / 98.49           &              80.67 / 3.14\\ 
                                                                                    & BIM &31.58 / 97.01           &30.84 / 97.26           &41.45 / 94.23          &37.16 / 98.25           &50.39 / 98.14         &       19.55 / 4.02 \\ 
                                                                                    & FGSM&0.04 / 86.59           &0.18 / 71.88           &83.52 / 99.17          &0.21 / 70.05          &81.86 / 93.61             &         83.48 / 29.12 \\
                                                                                    & MIM&35.80 / 95.34           &35.35 / 95.70            &78.70 / 98.13            &51.01 / 98.89        &61.71 / 98.41          &              43.35 / 3.55 \\
                                                                                    & FFGSM&0.00 / 95.07          &0.00 / 94.07            &40.53 / 98.52          &0.00 / 94.45          &99.02 / 99.17            &               99.02 / 5.10\\\midrule[1pt]
                                                            
        \multirow{3}{*}[-2.2ex]{\rotatebox{90}{\footnotesize Fashion \cite{xiao2017fashion}}}    & PGD &85.52 / 94.63            &84.30 / 92.01           &84.89 / 89.24           &84.55 / 91.47              &86.62 / 92.75    &         2.32 / 5.39\\ 
                                                                                    & BIM &87.00 / 93.96            &86.14 / 94.68              &86.54 / 91.95           &86.34 / 94.99              &87.54 / 94.01 &      1.40 / 3.04\\ 
                                                                                    & FGSM&64.82 / 90.68             &65.24 / 80.54            &90.50 / 96.35           &66.36 / 93.57              &76.82 / 95.47  &         25.68 / 15.81 \\
                                                                                    & MIM&86.15 / 93.11             &85.22 / 92.75            &85.65 / 95.12             &85.35 / 96.42              &87.01 / 95.12 &           1.79 / 3.67 \\
                                                                                    & FFGSM&57.28 / 80.06             &57.04 / 54.70          &88.48 / 95.98             &57.92 / 88.14              &87.81 / 96.95 &        31.44 / 42.25\\\bottomrule[1.5pt]

\end{tabular}
\label{tab:table2}
\end{table*}

\subsection{Experimental Results and Analysis}

\subsubsection{Defense Performance}

We compare the performance of our TRN with state-of-the-art defense methods including input transformation based methods \cite{guo2017countering,xu2017feature,xie2017mitigating,mustafa2019image,naseer2020self}, adversarial training \cite{madry2017towards}, and label smoothing \cite{warde201611}. We evaluate these defense methods against attack methods including FGSM \cite{goodfellow2014explaining}, BIM \cite{kurakin2016adversarial}, PGD \cite{madry2017towards}, MIM \cite{dong2018boosting} and FFGSM \cite{wong2020fast}. For adversarial training method and our method, we use the MIM attack method to generate adversarial images for training, and test on different attack methods. Follow the setting in \cite{yuan2020ensemble}, we set the iteration number as $10$ for iterative attack methods. For CIFAR10 and Fashion MNIST datasets, all attack methods are constrained in the perturbation budget $\epsilon = 8/255$ with each iteration step size as $1/255$. For SVHN dataset, the perturbation budget is set to $\epsilon = 12/255$ with a $3/255$ iteration step size following the setting in \cite{yuan2020ensemble}. In the experiment in Table \ref{tab:table1}, our TRN consists of 3 Fusion Block.

Table \ref{tab:table1} shows the classification accuracy of different defense methods under different attacks on three datasets. Our method outperforms other state-of-the-art defense methods under all five attacks on all three datasets. In particular, when under the MIM attack, our method outperforms other defenses by $30.72\%$, $35.48\%$, and $11.07\%$ on CIFAR10, SVHN, and Fashion MNIST, respectively. 

Our method can keep the performance of the target model on benign images, as validated on SVHN and Fashion MNIST (it only slightly degrades from 95.34\% to 94.37\% on CIFAR10). It can even improve the accuracy on benign images by $0.01\%$ on the SVHN dataset since our TRN can focus on the critical pixels that lead to the erroneous outputs with the guidance of gradient maps. We also observe that defended by our method, the model can perform better on adversarial images than on benign examples on the SVHN dataset (``Label Leaking Effect \cite{kurakin2016adversarial}''). To our experience, the Label Leaking Effect occurs only when the defense model has enough capacity to capture both the label-leaking and benign information. Thus, the results show our model has a larger capability to explore the patterns in adversarial examples.

\subsubsection{Generalizability across Attacks}

We also evaluate the generalizability across different attacks when the TRN is trained with one attack method. As Table \ref{tab:table2} shows, we use the attack methods in the first column to train our TRN and test on the attack methods in the first row. By comparing our method with adversarial training \cite{madry2017towards}, we can see that the adversarial training may fail to generalize to other attacks, but our method can generalize from one attack to another without significant degradation in performance. The results also demonstrates that our TRN obtains the most robust performance when training with the MIM attack. The largest performance degradation when training with the MIM attack is $4.96\%$ on CIFAR10.

\subsubsection{Robustness against Bypass Attacks}

Pre-processing defense methods that obfuscate gradients are vulnerable to the Backward Pass Differentiable Approximation attack (BPDA) \cite{athalye2018obfuscated}. BPDA can bypass the pre-processing defense by using an approximation network to estimate the gradients, though the gradient is unavailable in some non-linear pre-processing algorithms. We show that our defense method is resistant to BPDA.

Incorporating two streams in our TRN, the gradient map is hard to be attacked since the attacker will need to compute the gradients of the gradient map. In this case, we choose the image stream to estimate the gradients of inputs and further attack our TRN using PGD. As shown in Table \ref{tab:table3}, comparing to the performance of our defense under the PGD attack, the PGD-based BPDA attack only slightly degrades the accuracy. On the SVHN dataset, the performance only degrades the accuracy of TRN by $0.02\%$. This is because our defense contains gradient-stream and image-stream, and two streams complement each other to make our defense robust.

\begin{table}

\caption{The comparison of TRN's performance under the PGD attack and the BPDA attack. The gap between the performance of the PGD attack and the BPDA attack is also shown.}
\centering
\begin{tabular}{l c c c }
\toprule[1.5pt]
Attacks   & CIFAR10 & SVHN & Fashion MNIST \\ \midrule[1pt]
PGD \cite{madry2017towards}  & 89.72  & 97.54 &  94.63\\ 
BPDA \cite{athalye2018obfuscated} & 87.86  &  97.52 & 93.75  \\ \midrule[1pt]
gap & 1.86 & 0.02  & 0.88 \\
\bottomrule[1.5pt]
\end{tabular}
\label{tab:table3}
\end{table}

\section{Ablation Study}

\subsection{Is two-stream better than one-stream?}

\begin{figure*}
    \centering
    \includegraphics[scale=0.5]{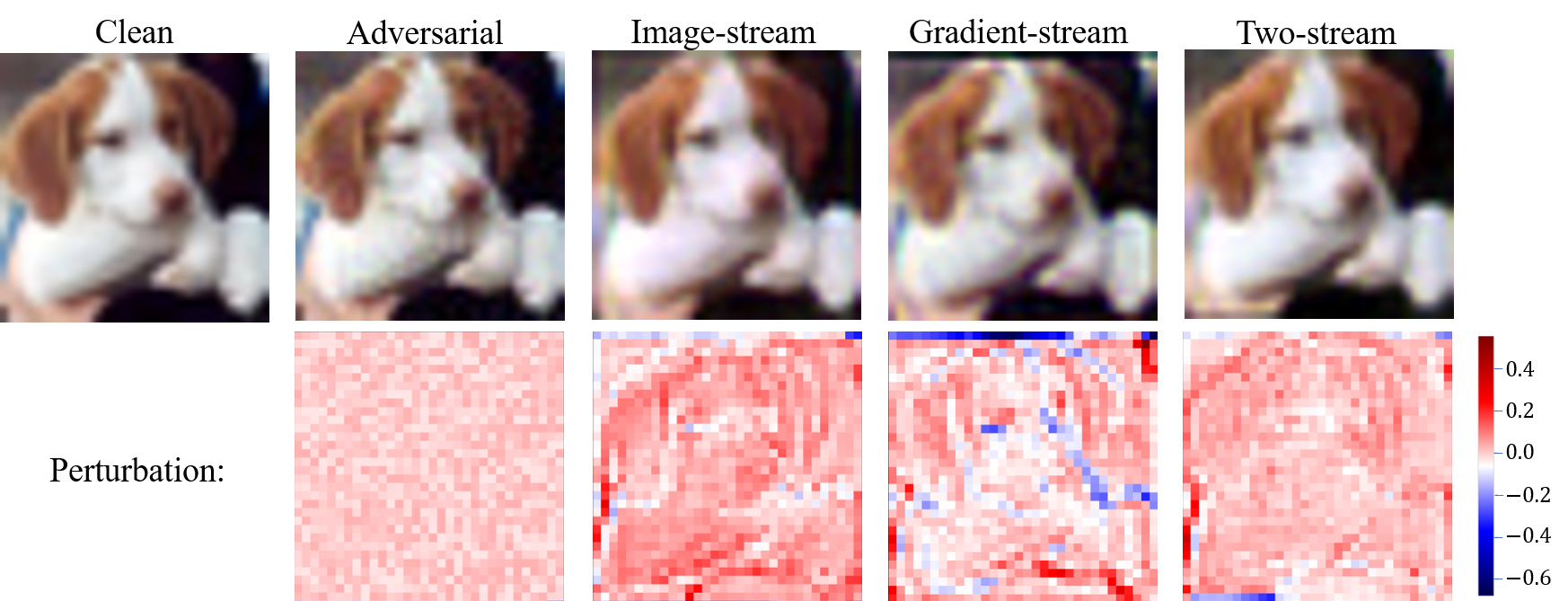}
    \caption{Examples of a pre-processed image with two streams and single stream. We show the images and the perturbation of restored or perturbed images. We can see that relying on a single stream, the restored images contain high-vibration perturbation, while with gradient and image streams, the network can restore the images more precisely (flatten and small perturbation). }
    \label{fig:Ablation}
\end{figure*}

\begin{figure}
    \centering
    \includegraphics[scale=0.45]{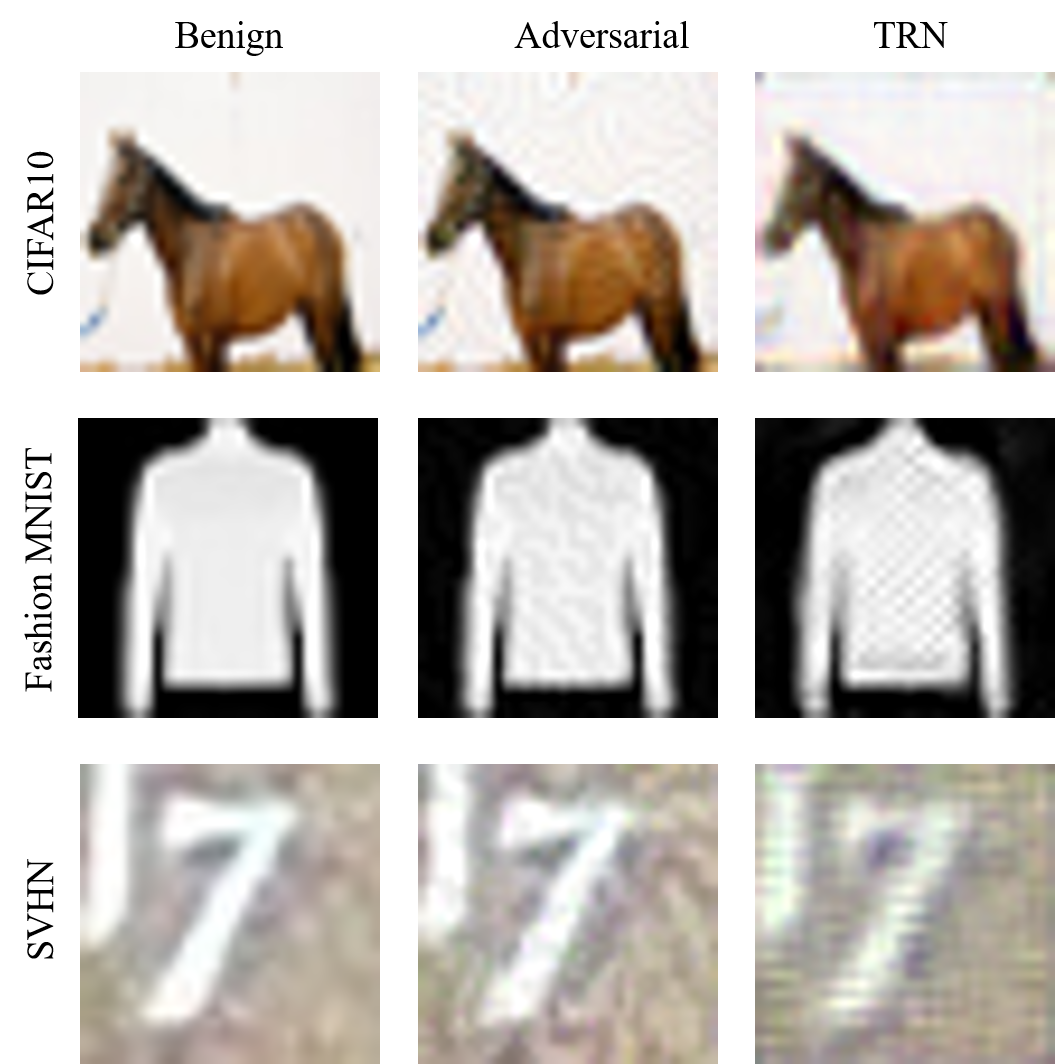}
    \caption{Examples of restored images on CIFAR10, SVHN and Fashion MNIST.}
    \label{fig: Discussion}\vspace{-0.2in}
\end{figure}

\begin{table}

\caption{The comparison of different streams against PGD on CIFAR10. We show the results (classification accuracy after defense) against PGD attack from 10 iterations to 40 iterations. }
\centering
\begin{tabular}{l c c c c}
\toprule[1.5pt]
Methods         & 10 iter.  & 20 iter.      & 30 iter.      & 40 iter. \\ \midrule[1pt]
No defense      & 0.01      & 0.00          & 0.00          & 0.00      \\ 
Gradient Stream & 87.09     &  85.78        & 85.92              & 85.34            \\ 
Image Stream    & 87.41     &   87.60       & 87.34              & 86.80          \\ \midrule[1pt]
Two-stream      & \textbf{89.72}     &   \textbf{89.19 }      &  \textbf{88.86}             & \textbf{88.18}           \\ \bottomrule[1.5pt]
\end{tabular}
\label{tab:table4}\vspace{-0.1in}
\end{table}

In this section, we compare our two-stream restoration network with one-stream restoration network. We implement the gradient stream restoration network by removing the convolution layers in the image stream except for the last layer. This means we only use the gradient maps to estimate the benign perturbation and add the benign perturbation on adversarial images without feeding adversarial images to the network. In this case, we do not extract any features from the adversarial images. To implement the image stream restoration network, we simply remove the gradient stream. 

In Table \ref{tab:table4}, we compare the complete TRN with restoration networks with image stream or gradient stream, respectively. From the results of gradient stream defense, we can see that by only relying on the gradient maps, we can still recover the adversarial images to defend against strong attacks like PGD. This means that the gradient maps contain enough information to defend against gradient-based attacks. Comparing the results of the two-stream method and one-stream methods, we can see that the two-stream manner reaches the best accuracy on PGD-attacked images. It indicates that the TRN can explore the complementary information in two streams. Since the gradient of the adversarial images measures how each pixel of adversarial images contribute to the erroneous outputs, the restoration network can focus on the critical pixels with the guidance of the gradient stream (see Figure \ref{fig:Ablation}). Thus, our TRN provides stronger defense to the target model.

\subsection{Scalability}

\begin{table}

\caption{The performance of our method (Classification accuracy after defense) with different number of Fusion Block against PGD attack on CIFAR10, SVHN and Fashion MNIST.}
\centering
\begin{tabular}{c c  c  c }
\toprule[1.5pt]
\small Block Number    &\small  CIFAR10 &SVHN & Fashion MNIST  \\ \midrule[1pt]
No defense & 0.01  & 0.16  &0.00   \\ 
1 &  89.90 &  97.18 & 94.01\\ 
2 &  90.05 &  97.68 & 94.04 \\ 
3 & \textbf{90.29}  & 97.30  & 94.30 \\ 
4 & 90.13  &  96.79 &94.22    \\ 
5 & 89.92  &  96.31 &\textbf{94.65}\\ 
6 & 90.04  &  97.58 &93.46\\
7 & 90.12  &  \textbf{97.84} &94.34\\
8 & 90.06  &  96.18 &92.93\\
9 & 90.11  &  92.15 &93.38\\ \bottomrule[1.5pt]
\end{tabular}
\label{tab:table5}\vspace{-0.1in}
\end{table}

With both image and gradient streams as inputs, our FB refines the gradient map and restores the adversarial images progressively. The residual connections also enable the outputs to maintain the raw information in inputs. Then, our FB can be simply cascaded to construct a deeper network. In Table \ref{tab:table5}, we evaluate our TRN with different Fusion Block numbers against the PGD attack. Table \ref{tab:table5} shows that TRN achieves the best accuracy on adversarial images with 3, 7 and 5 Fusion Blocks on CIFAR10, SVHN and Fashion MNIST, respectively. Due to the scalability of our method, we can change the number of FB to obtain the optimal defense on different datasets which need different model capabilities. Note that reducing the FB numbers to one only leads to a minor degradation from the performance with optimal block numbers. Thus, our method can approximate the best performance by training a small-scale TRN when the computational resources are limited.

\subsection{Discussion}

\begin{figure}
    \centering
    \includegraphics[scale=0.38]{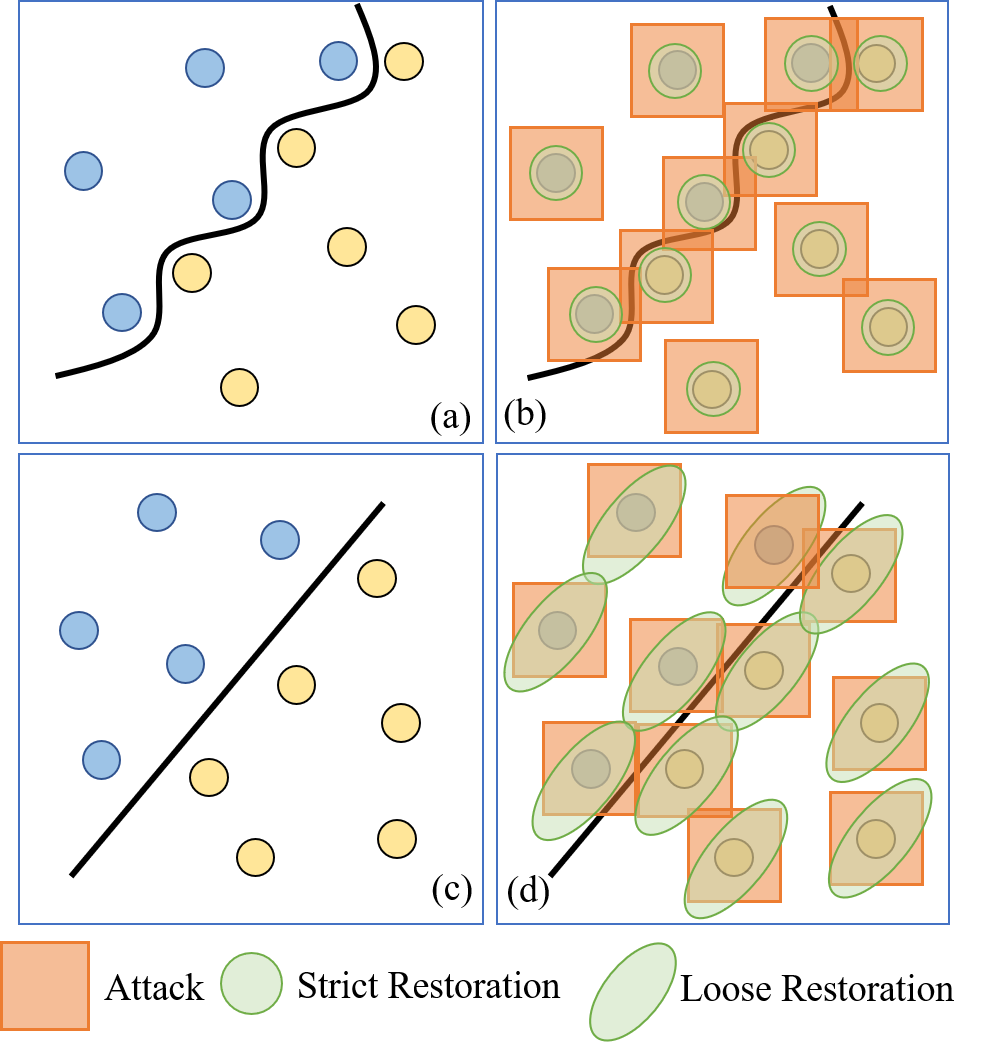}
    \caption{A conceptual illustration of benign, adversarial, and restored examples. In (a) and (c), a set of points can be separated by the simple (Linear) or complicated decision boundary. In (b), the decision boundary does not separate the $\ell_\infty$-balls around the data points, but after the strict restoration, the decision boundary can successfully separate the restored data points. In (d), to defend the target model with a simple decision boundary, the loose restoration is sufficient, but the loose restoration can not defend target models with a complicated decision boundary.}
    \vspace{-0.2in}
    \label{fig:Illustration}
\end{figure}

We visualize the image restored by our method under the MIM attack (see Figure \ref{fig: Discussion}). The examples of CIFAR10 illustrate that our method can clean the adversarial pattern on images. However, on the SVHN and Fashion MNIST, we surprisingly observe that our TRN add some grid structures to defend the attack. This inspires us to think about the nature of restoration. As shown in Figure \ref{fig:Illustration}, since the data distribution is simpler in SVHN and Fashion MNIST than CIFAR10, it is sufficient for TRN to defend these datasets by adding some robust grid patterns (``loose restoration''). Noticeably, these robust grid patterns will revert the accuracy on adversarial examples without affecting the benign examples since these patterns are generated along the decision boundary (see (d) in Figure \ref{fig:Illustration}). However, for a dataset with complicated distributions like CIFAR10, TRN needs to learn a strict restoration to revert the performance of the target model. Thus, Figure \ref{fig: Discussion} shows that the restored examples on CIFAR10 will be closer to the benign images.


\section{Conclusion}

To our best knowledge, we propose the first defense method that incorporates the gradients into pre-processing for defending against gradient-based attacks. A novel Two-stream Restoration Network is proposed to jointly use the estimated gradient maps and adversarial images to restore adversarial images. A Fusion Block is also proposed to explore and fuse the information in the gradient space and image space. Our defense provides a new insight to input transfromation based defenses in which the estimated gradient maps may contain complementary information besides the adversarial images. It successfully defends against the state-of-the-art attacks on three representative datasets.

{\footnotesize
\bibliographystyle{ieee_fullname}
\bibliography{main}
}

\end{document}